\def\year{2019}\relax
\documentclass[letterpaper]{article} %DO NOT CHANGE THIS
\usepackage{aaai19}  %Required
\usepackage{times}  %Required
\usepackage{helvet}  %Required
\usepackage{courier}  %Required
\usepackage{url}  %Required
\usepackage{graphicx}  %Required
\usepackage{amsmath}
\usepackage{fancyhdr,graphicx,amssymb}
\usepackage[ruled,vlined]{algorithm2e}
\usepackage{multirow}
\newcommand{\newref}[1]{Eq.\eqref{#1}}
\newcommand{\bx}{\mathbf{x}}
\newcommand{\by}{\mathbf{y}}
\newcommand{\righttheta}{\theta_{\bx\rightarrow \by}}
\newcommand{\lefttheta}{\theta_{\bx\leftarrow \by}}
\newcommand{\citet}[1]{\citeauthor{#1}~\shortcite{#1}}

\begin{document}
% The file aaai.sty is the style file for AAAI Press 
% proceedings, working notes, and technical reports.
%
\title{Unsupervised Neural Machine Translation with \\
SMT as Posterior Regularization}
\author{Shuo Ren$^\dag$\thanks{The first two authors contributed equally to this work. This work is supported in part by NSFC U1636210 and 61421003, and Shenzhen Institute of Computing Sciences.}, Zhirui Zhang$^\ddag$, Shujie Liu$^\S$, Ming Zhou$^\S$, Shuai Ma$^\dag$ \\
$^\dag$SKLSDE Lab, Beihang University \ $^\dag$Beijing Advanced Innovation Center for Big Data and Brain Computing, China\\
$^\ddag$University of Science and Technology of China, Hefei, China \ 
$^\S$Microsoft Research Asia \\
$^\dag$\{shuoren,mashuai\}@buaa.edu.cn\ $^\ddag$zrustc11@gmail.com \ 
$^{\S}$\{shujliu,mingzhou\}@microsoft.com \\
}
\maketitle
\begin{abstract}
Without real bilingual corpus available, unsupervised Neural Machine Translation (NMT) typically requires pseudo parallel data generated with the back-translation method for the model training. 
However, due to weak supervision, the pseudo data inevitably contain noises and errors that will be accumulated and reinforced in the subsequent training process, leading to bad translation performance. 
To address this issue, we introduce phrase based Statistic Machine Translation (SMT) models which are robust to noisy data, as posterior regularizations to guide the training of unsupervised NMT models in the iterative back-translation process.
Our method starts from SMT models built with pre-trained language models and word-level translation tables inferred from cross-lingual embeddings. Then SMT and NMT models are optimized jointly and boost each other incrementally in a unified EM framework. 
In this way, (1) the negative effect caused by errors in the iterative back-translation process can be alleviated timely by SMT filtering noises from its phrase tables; meanwhile, (2) NMT can compensate for the deficiency of fluency inherent in SMT.
Experiments conducted on \emph{en}-\emph{fr} and \emph{en}-\emph{de} translation tasks show that our method outperforms the strong baseline and achieves new state-of-the-art unsupervised machine translation performance.
\end{abstract}

\section{Introduction}
\label{sec_introduction}
Recent years have witnessed the rise and success of Neural Machine Translation (NMT) ~\cite{sutskever2014sequence,bahdanau2014neural,luong2015effective,wu2016google,vaswani2017attention,hassan2018achieving}. However, NMT relies heavily on large in-domain parallel data, resulting in poor performance on low-resource language pairs \cite{koehn2017six}. 
For some low-resource pairs without any bilingual corpus, how to train NMT models with only a monolingual corpus is a popular and interesting topic.

\begin{figure}[!htb]
\centering
\includegraphics[width=\linewidth]
{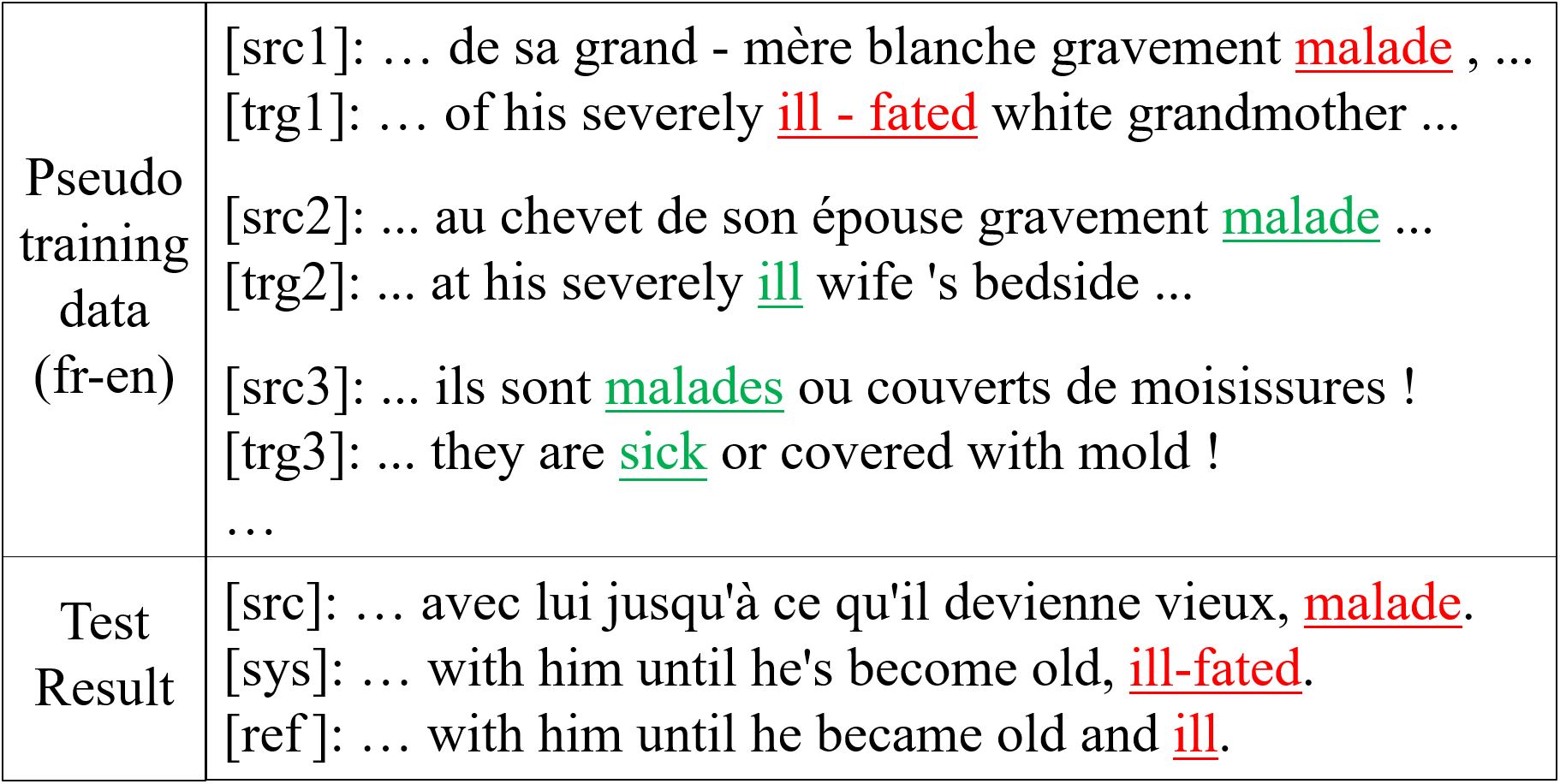}
\caption{The effect of noisy training data. The first training sample contains the noise (``malade" in French means ``ill", not ``ill-fated"), leading to the wrong test result (sys).}
\label{fig:example}
\end{figure}

Existing methods for unsupervised machine translation \cite{artetxe2017unsupervised,lample2017unsupervised,yang2018unsupervised,lample2018phrase} are mainly the modifications of encoder-decoder schema. In their work, source sentences in two languages are mapped into the same latent space with a shared encoder, which is expected to be the internal information representation irrelevant to the languages themselves. From that target sentences are generated by a shared or different decoders. Some of them also use denoising auto-encoders \cite{vincent2010stacked} and adversarial training. Despite the differences in structures and training methods, they reach a consensus to use the pseudo parallel data generated iteratively with the back-translation method \cite{sennrich2016improving,zhang2018joint} to train their unsupervised NMT models,
%\cite{sennrich2016improving,zhang2018joint} 
%thus the unsupervised problem can be converted into the supervised one, 
i.e. they use monolingual data in the target language and a target-to-source translation model to generate source sentences, then use the pseudo parallel data of generated sources and real targets to train the source-to-target model, and vice versa.

However, since the pseudo data are generated by unsupervised models, random errors and noises are inevitably introduced, such as redundant or unaligned words deviating from the meaning of source sentences. Due to the lack of supervision, those infrequent errors will be accumulated and reinforced by NMT models into frequent patterns during the training iterations, leading to bad translation performance.
For instance in Figure \ref{fig:example}, the French word ``malade" is mistakenly translated into the English word ``ill-fated" in the first training sample. With strong abilities to identify and memorize patterns, NMT models mistakenly translate this word into ``ill-fated" when ``old" (similar to ``grandmother" in the first training sample) occurs in the test. 
Even so, there are also many good translation patterns (such as ``malade" $\rightarrow$ ``ill" or ``sick" in the second and third training samples), which could have been extracted in time to guide the NMT models into the correct training direction. The extraction and guidance can be well carried out by Statistical Machine Translation (SMT).
As is pointed out by \citeauthor{khayrallah2018impact} \shortcite{khayrallah2018impact}, SMT performs better than NMT in tackling noisy data by constructing a strong phrase table with good and frequent translation patterns and filtering out infrequent errors and noises.
This gives the motivation that if we incorporate SMT in the training process, unsupervised NMT could benefit from the robustness of SMT to noisy data.

In this paper, we propose to leverage SMT to denoise and guide the training of unsupervised NMT models in the iterative back-translation process. Different from previous work \cite{he2016improved,tang2016neural,wang2017neural} introducing SMT into NMT by changing model structures in supervised scenarios, we adopt the framework of posterior regularization \cite{ganchev2010posterior} to leave model structures unchanged. 
Our method starts from initial SMT models built with pre-trained language models and word-level translation tables inferred from cross-lingual embeddings. Then SMT models and NMT models are trained jointly in a unified Expectation Maximization (EM) training framework. 
In each iteration, as desired distributions, SMT models are expected to correct NMT models timely with denoised pseudo data generated in a constrained search space of reliable translation patterns. Based on that, enhanced NMT models can generate better pseudo data for SMT to extract phrases of higher quality, so that they can benefit from each other incrementally.
In this way, infrequent errors in NMT models can be eliminated with the constraints exerted by SMT features, while NMT can compensate for the deficiency in smoothness inherent in SMT models.
Experiments conducted on en-fr and en-de translation tasks show that our method significantly outperforms the strong baseline \cite{lample2018phrase} and achieves the new state-of-the-art translation performance in unsupervised machine translation.

\section{Background}
%In this section, we will post the background of this paper. We first introduce the current neural machine translation method in the first subsection. And Phrase-based statistic machine translation system will be introduced in the next subsection. Finally, we will introduce posterior regularization.
\subsection{Neural Machine Translation}
Given a source sentence $\bx=(x_1,x_2,...,x_l)$ and a target one $\by=(y_1,y_2,...,y_m)$, Neural Machine Translation (NMT) directly models the word-level translation probability with parameters $\theta$ as:
\begin{equation}
p(y_i|\bx,\by_{<i};\theta)=\textup{softmax}(g(\mathbf{h}_{y_i},\mathbf{h}_{y<i}, \mathbf{c}_{i};\theta))
\label{NMT_softmax}
\end{equation}
in which $g(\cdot)$ denotes a non-linear function extracting features to predict the target word $y_i$ from the decoder states ($\mathbf{h}_{y_i}$ and $\mathbf{h}_{y_{<i}}$)  
%of the current word $y_i$ and previous translated words $y_{<i}$ respectively, 
and the context vector $\mathbf{c}_{i}$ calculated with the encoder and attention mechanism. Then the sentence-level translation probability $p(\by|\bx;\theta)$ is calculated by $p(\by|\bx;\theta)=\prod_{i=1}^{m}p(y_i|\bx,\by_{<i};\theta)$. As for training, given a parallel corpus $\{(\bx_{n},\by_{n})\}_{n=1}^{N}$, the objective function is to maximize $\log p(\by_{n}|\bx_{n};\theta)$ over the whole training set.

\subsection{Phrase-based Statistic Machine Translation}
\label{PBSMT}
%Based on the noisy channel model, Phrase-based Statistic Machine Translation (PBSMT) uses Bayes rules to reformulate the translation probability for translating a source sentence $\bx$ into a target sentence $\by$ ~\cite{koehn2003statistical} as follows:
%\begin{equation}
%\arg\max_{\by}p(\by|\bx)=\arg\max_{\by}p(\bx|\by)p(\by)
%\label{SMT_noise-channel-model}
%\end{equation}
%where $p(\bx|\by)$ is a translation model derived from a phrase table, and $p(\by)$ is a language model of the target language. 

%Apart from the noisy channel based model, ~\citet{och2002discriminative} proposed the log-linear model for statistic machine translation, in which translation probability from $\bx$ to $\by$ is formulated as:
The current approach of Statistic Machine Translation (SMT) is typically based on the log-linear model proposed by \citeauthor{och2002discriminative} \shortcite{och2002discriminative}. According to it, the translation probability from sentence $\bx$ to sentence $\by$ is formulated as:
\begin{equation}
p(\by|\bx;\lambda_1^M)
=\frac{\exp{[\sum_{m=1}^M\lambda_{m}h_{m}(\bx,\by)]}}
{\sum_{\mathbf{\tilde{\by}}}\exp{[\sum_{m=1}^M\lambda_{m}h_{m}(\bx,\tilde{\by})]}}
\label{SMT_log-linear-model}
\end{equation}
where $h_{m}(\bx,\by)=\log \phi_m(\bx,\by)$ denotes the m$^{th}$ feature.
%, and $\lambda_{m}$ is the weight apportioned to it. 

In phrase based SMT (PBSMT) \cite{koehn2003statistical}, the sentence pair is segmented into a sequence of phrases $\bar{\bx}_1^I$ and $\bar{\by}_1^J$, where $I$ and $J$ are the counts of phrases. During training, given a bilingual corpus, PBSMT first infers word alignment, based on which phrase pairs are derived and stored in the phrase table, as well as translation probabilities. 
% between phrases in different languages. 
Other features such as a distortion model 
%$d(i,j)$, which means the probability that a phrase at position $i$ is translated into a phrase at position $j$, 
can also be learned with the extracted phrase pairs.
The feature weights $\lambda_1^M$ can be optimized by MERT \cite{och2003minimum} with a validation set. 
During decoding, PBSMT generates translation candidates $\tilde{\by}$ bottom up via the CKY algorithm, ranked with scores given by the log-linear model in \newref{SMT_log-linear-model}. 

\subsection{Posterior Regularization}
Posterior regularization ~\cite{ganchev2010posterior} is a framework for structured, weakly supervised learning, which incorporates indirect supervision from a desired distribution $q(\by)$ via constraints on posterior distribution $p(\by|\bx_{n};\theta)$ imposed by a Kullback-Leible (KL) divergence as follows:
\begin{equation}
\begin{aligned}
F(q;\theta) = \mathcal{L}(\theta)-\sum_{n=1}^{N}\min_{q \in Q}\mathbf{KL}(q(\by)||p(\by|\bx_{n};\theta)) \\
\end{aligned}\label{PR_Likelihood}
\end{equation}
where $\mathcal{L}(\theta)$ is the original likelihood of model $p(\by|\bx;\mathbf{\theta})$, and $Q$ is a constraint posterior set satisfying:
\begin{equation}
Q = \{q(\by) : \mathbf{E}_{q}[\phi(\bx,\by)]\leq \mathbf{b}\}
\label{Q_set}
\end{equation}
%in which the expectations of constraints features $\mathbf{\phi}(\bx,\by)$ are bounded by the hyper-parameter $\mathbf{b}$.
in which constraints features $\mathbf{\phi}(\bx,\by)$ are bounded by $\mathbf{b}$.

To maximize $F(q;\mathbf{\theta})$, \citeauthor{ganchev2010posterior} \shortcite{ganchev2010posterior} propose an EM framework \cite{mclachlan2007algorithm} as:
\begin{equation}
\begin{aligned}
E: q^{t+1} 
%&= \mathop{\arg\max}_{q \in Q}F(q;\theta) \\
&= \mathop{\arg\min}_{q \in Q}\mathbf{KL}(q(\by)||p(\by|\bx_{n};\theta^{t}))\\
M: {\theta}^{t+1} 
%&= \mathop{\arg\max}_{\theta}F(q^{t+1};\theta) \\
&= \mathop{\arg\max}_{\theta}\mathcal{L}(\theta) + \mathbf{E}_{q^{t+1}}[\log p(\by|\bx_{n};\theta)]
\end{aligned}
\label{PR_EM}
\end{equation}

%Based on our motivation proposed before, we want to improve the NMT model denoted by $p(\by|\bx_{n};\mathbf{\theta}))$ with SMT as posterior regularization. 
However, there may be a problem as pointed out by \citeauthor{zhang2017prior} \shortcite{zhang2017prior} that it is hard to set a reasonable bound $\mathbf{b}$ if we directly apply posterior regularization to NMT. To solve this problem, we follow their practice of representing the desired distribution $q(\by)$ as the log-linear model described in \newref{SMT_log-linear-model}. In this way, SMT models directly act as the posterior regularization to constrain NMT models $p(\by|\bx_{n};\theta^{t})$.
%\textcolor{red}{There \textbf{may be a problem} as pointed in \cite{zhang2017prior}  that it is hard to directly apply posterior regularization to neural machine translation because the effective bound $\mathbf{b}$ is difficult to specify. However, in our method, this problem can be solved naturally because we can specify this bound as the feature value of trained SMT systems.}

\section{Method}
\subsection{Overview}
\label{sec_overview}

\begin{figure*}[!htb]
\centering
\includegraphics[height=6.3cm]{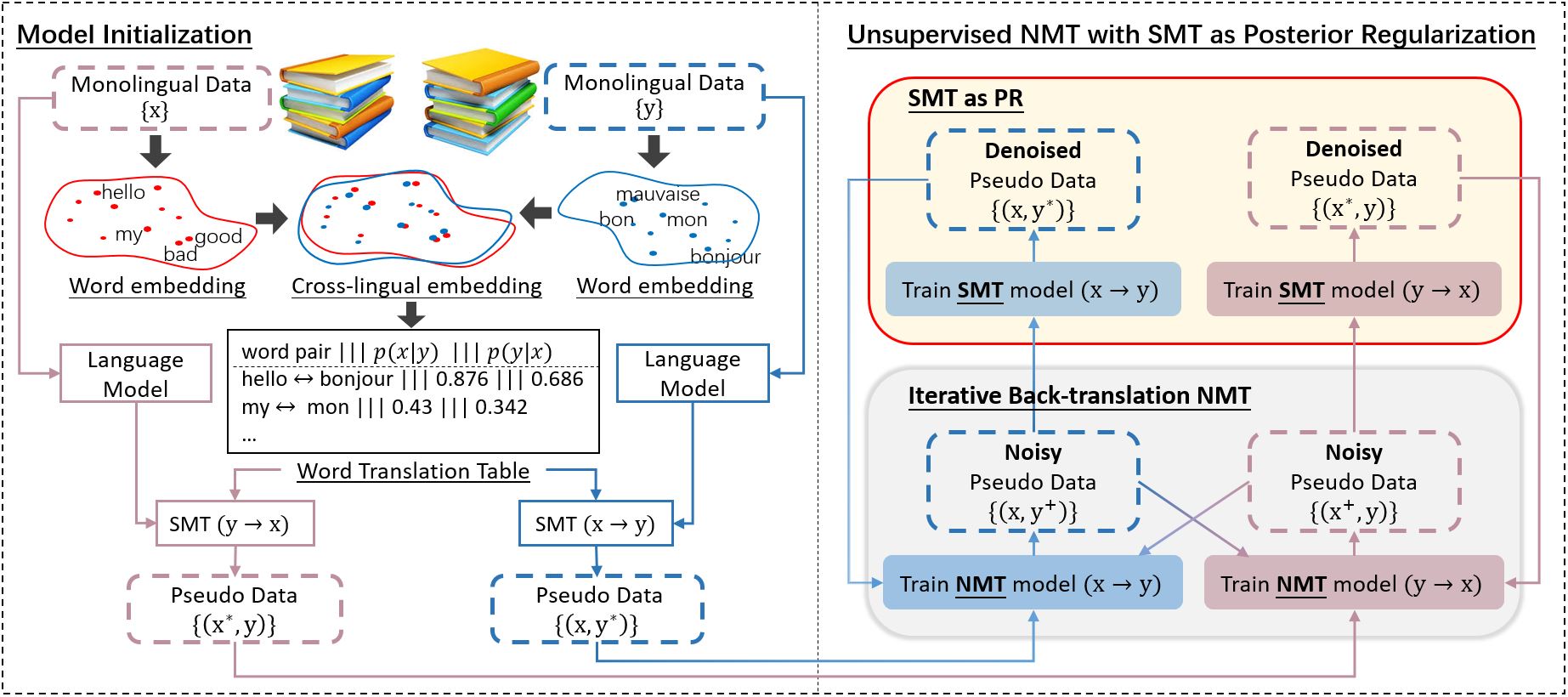}
\caption{Method overview. The whole procedure mainly consists of two parts as the left and the right.}
\label{fig:overview}
\end{figure*}

Due to the lack of supervision, noises and infrequent errors in the pseudo data generated by unsupervised NMT models will be accumulated and reinforced in the iterative back-translation process (shown in the shadow area in Figure \ref{fig:overview}). To address this issue, we introduce SMT as posterior regularization (the red frame above that) to denoise and guide the training of NMT, thus the noises being eliminated timely.
%The whole procedure of our method mainly consists of two parts as shown in Figure \ref{fig:overview}.
%We will introduce the whole procedure of our method in Figure \ref{fig:overview} as follows.

% Our approach mainly consists of model initialization and leveraging SMT as posterior regularization parts.
The whole procedure of our method mainly consists of two parts shown in the left and right of Figure \ref{fig:overview}.
Given a language pair X-Y, for \textbf{\underline{model initialization}}, we build two initial SMT models with language models pre-trained using monolingual data, and word translation tables inferred from cross-lingual embeddings according to the approach in \ref{sec_initialization}. 
Then the initial SMT models will generate pseudo data to warm up two NMT models. Note that the NMT models are trained using not only the pseudo data generated by SMT models, but those generated by reverse NMT models with the \textbf{\underline{iterative back-translation}} method. After that, the NMT-generated pseudo data are fed to SMT models. As \textbf{\underline{posterior regularization (PR)}}, SMT models timely filter out noises and infrequent errors by constructing strong phrase tables with good and frequent translation patterns, and then generate denoised pseudo data to guide the subsequent NMT training. Benefiting from that, NMT then produces better pseudo data for SMT to extract phrases of higher quality, meanwhile compensating for the deficiency in smoothness inherent in SMT via back-translation. Those two steps are unified in the EM framework described in \ref{sec_objective}, where NMT and SMT models are trained jointly and boost each other incrementally until final convergence.
%Previous methods \cite{artetxe2017unsupervised,lample2017unsupervised,yang2018unsupervised,lample2018phrase} mainly focus on the shadow area in Figure \ref{fig:overview}, i.e. iterative back-translation. However, without any regularization method, the noises or infrequent errors in the pseudo data generated by NMT models will be repeated and reinforced during the back-translation iterations, especially in the fully unsupervised scenario without any real bilingual data. To address this issue, in our method, SMT is introduced as a posterior regularization and plays the role of denoising for NMT models, as shown in the red frame in Figure \ref{fig:overview}, since SMT could better tackle with the noisy data \cite{khayrallah2018impact} by constructing strong phrase tables with good and frequent translation patterns and filtering out the detrimental information.  
%In the following subsections, we will first give the details of model initialization as shown in the left part of Figure \ref{fig:overview}. To illustrate the right part of Figure \ref{fig:overview}, We will derive the optimization objective and give the EM steps for training in Section \ref{sec_objective}. Finally, we will combine the whole procedure together into an unified algorithm shown in Section \ref{sec_training}.

\subsection{Initialization}
\label{sec_initialization}
Our initial SMT models are built with word-based phrase tables and two pre-trained language models via Moses\footnote{https://github.com/moses-smt/mosesdecoder}. For the word translation table, we first train word embeddings using monolingual corpora for two languages respectively. Based on that, we adopt the method proposed by Artetxe et al. \shortcite{artetxe2018aaai} to obtain respective cross-lingual embeddings $\{e_{x_i}\}_{i=1}^S$ and $\{e_{y_j}\}_{j=1}^T$, where $S$ and $T$ are vocabulary sizes. Then the word translation probability from word $x_i$ to $y_j$ is:
\begin{equation}
\begin{aligned}
p(y_j|x_i)=\frac{\exp{[\lambda\cos({e_{x_i},e_{y_j}}})]}{\sum_k\exp{[\lambda\cos({e_{x_i},e_{y_k}}})]}
\end{aligned}\label{word_translation_prob}
\end{equation}
where $\lambda$ is a hyper-parameter to control the peakiness of the distribution. The calculation of $p(x_i|y_j)$ is similar to \newref{word_translation_prob}. Based on the above, we choose top-$k$ translation candidates for each word in our initial phrase table. We only use two features in our initial phrase tables, i.e. translation probabilities and inverse translation probabilities.
%We choose $\lambda=5$ in our experiments. And we select top-10 translation candidates for each word and set the vocabulary size for each language as 100,000. 

\subsection{Unsupervised NMT with SMT as PR}
\label{sec_objective}
As is mentioned in \ref{sec_overview}, SMT plays a role in denoising and is leveraged as posterior regularization for NMT. Therefore, we replace the posterior regularization term $q(\by)$ in \newref{PR_Likelihood} with the SMT models ($\bx\rightarrow \by$) and ($\by\rightarrow \bx$) in Figure \ref{fig:overview}, which will be denoted by $\overrightarrow{p_s}(\by|\bx)$ and $\overleftarrow{p_s}(\bx|\by)$. 
%Note that in our SMT models, we use six features including language model, word penalty, phrase penalty, forward translation, backward translation, and unknown word penalty, which is also the default setting of Moses \footnote{https://github.com/moses-smt/mosesdecoder} implementation of PBSMT systems. 
By the way, the NMT models ($\bx\rightarrow \by$) and ($\by\rightarrow \bx$) will be denoted by $\overrightarrow{p_n}(\by|\bx;\righttheta)$ and $\overleftarrow{p_n}(\bx|\by;\lefttheta)$, where $\righttheta$ and $\lefttheta$ are parameters. Then,
% following \cite{ganchev2010posterior} and \cite{zhang2017prior}, 
given monolingual corpora $\{\bx_i\}_{i=1}^M$ and $\{\by_j\}_{j=1}^N$, we formulate the training objective as:

\begin{equation}
\begin{aligned}
& \mathcal{J}(\righttheta,\lefttheta, \overrightarrow{p_s}, \overleftarrow{p_s}) = \bar{\mathcal{L}}(\righttheta,\lefttheta)\\
& -\sum_{i=1}^{M}\min_{\overrightarrow{p_s}}\mathbf{KL}(\overrightarrow{p_s}(\by|\bx_i)||\overrightarrow{p_n}(\by|\bx_i;\righttheta)) \\
& -\sum_{j=1}^{N}\min_{\overleftarrow{p_s}}\mathbf{KL}(\overleftarrow{p_s}(\bx|\by_j)||\overleftarrow{p_n}(\bx|\by_j;\lefttheta)) 
\end{aligned}\label{Training_objective}
\end{equation}
where $\bar{\mathcal{L}}(\righttheta,\lefttheta)$ corresponds to the training objective of iterative back-translation for NMT models, which is
\begin{equation}
\begin{aligned}
& \bar{\mathcal{L}}(\righttheta,\lefttheta)\\
= &\sum_{i=1}^{M}\mathbf{E}_{\by\sim \overrightarrow{p_n}(\by|\bx_i;\righttheta)}[\log \overleftarrow{p_n}(\bx_i|\by;\lefttheta)]\\
+&\sum_{j=1}^{N}\mathbf{E}_{\bx\sim \overleftarrow{p_n}(\bx|\by_j;\lefttheta)}[\log \overrightarrow{p_n}(\by_j|\bx;\righttheta)]\\
\end{aligned}
\label{Unsupervised_NMT_goal}
\end{equation}
and two Kullback-Leibler divergence (KL) terms denote the posterior regularizations for two NMT models respectively. 

Based on that, the training processes of iterative back-translation for NMT and SMT models as posterior regularization are unified into a single objective $\mathcal{J}$. Then, we modulate the EM algorithm in \newref{PR_EM} to optimize it as follows: 

\begin{equation}
\begin{aligned}
E: \overleftarrow{p_s}^{t+1} 
&= \mathop{\arg\max}_{\overleftarrow{p_s}}\mathcal{J}(\righttheta,\lefttheta,\overrightarrow{p_s}, \overleftarrow{p_s}) \\
&= \mathop{\arg\min}_{\overleftarrow{p_s}}\mathbf{KL}(\overleftarrow{p_s}(\bx|\by_{j})||\overleftarrow{p_n}(\bx|\by_{j};\lefttheta^{t}))\\
\overrightarrow{p_s}^{t+1} 
&= \mathop{\arg\max}_{\overrightarrow{p_s}}\mathcal{J}(\righttheta,\lefttheta,\overrightarrow{p_s}, \overleftarrow{p_s}) \\
&= \mathop{\arg\min}_{\overrightarrow{p_s}}\mathbf{KL}(\overrightarrow{p_s}(\by|\bx_{i})||\overrightarrow{p_n}(\by|\bx_{i};\righttheta^{t}))\\
%\end{aligned}
%\label{E-step}
%\end{equation}
%\begin{equation}
%\begin{aligned}
M: \lefttheta^{t+1} 
&= \mathop{\arg\max}_{\lefttheta}\mathcal{J}(\righttheta,\lefttheta,\overrightarrow{p_s}, \overleftarrow{p_s}) \\
&= \mathop{\arg\max}_{\lefttheta}\{\mathbf{E}_{\overleftarrow{p_s}^{t+1}}[\log \overleftarrow{p_n}(\bx|\by_j;\lefttheta)]\\
&+\mathbf{E}_{\overrightarrow{p_n}(\by|\bx_i;\righttheta^{t})}[\log \overleftarrow{p_n}(\bx_i|\by;\lefttheta)]\}\\
\righttheta^{t+1} 
&= \mathop{\arg\max}_{\righttheta}\mathcal{J}(\righttheta,\lefttheta,\overrightarrow{p_s}, \overleftarrow{p_s}) \\
&= \mathop{\arg\max}_{\righttheta}\mathbf{E}_{\overrightarrow{p_s}^{t+1}}[\log \overrightarrow{p_n}(\by|\bx_i;\righttheta)]\\
&+\mathbf{E}_{\overleftarrow{p_n}(\bx|\by_j;\lefttheta^{t})}[\log \overrightarrow{p_n}(\by_j|\bx;\righttheta)]
\end{aligned}
\label{EM}
\end{equation}

Briefly speaking, in the E-step, we optimize the desired distributions represented by SMT to minimize the KL distance between SMT models and NMT models. In the M-step, we optimize NMT models using the pseudo data generated by SMT models and the corresponding reverse NMT models to fit the desired distributions and meanwhile perform back-translation iterations. We will give the specific equation for updating parameters in \ref{sec_training}. 

\subsection{Training Algorithm}
\label{sec_training}
We combine the model initialization and the whole training procedure into Algorithm \ref{alg:1} as follows.

\begin{algorithm}[!htb]
\small
\KwIn{Monolingual data $X = \{\bx_{i}\}_{i=1}^{M}$ and $Y=\{\by_{j}\}_{j=1}^{N}$}
\KwOut{Parameters of two NMT models: $\righttheta$, $\lefttheta$}
		\nl Train language models $l_\bx$ and $l_\by$ using $X$ and $Y$\\
		\nl Infer word translation tables $t_{\bx\by}$ and $t_{\by\bx}$ as in \ref{sec_initialization} \\ 
        \nl $t:=0$\\
        \While {not convergence}{
        \nl Sample data $\{\bx_t\} \in X$ and $\{\by_t\} \in Y$\\
        \nl \textbf{$//$ E-step:}\\
        \If {$t=0$}{
        \nl Initialize $\overrightarrow{p_s}^0$ and $\overleftarrow{p_s}^0$ using $l_\bx$, $l_\by$, $t_{\bx\by}$ and $t_{\by\bx}$}
        \Else{
        \nl Generate pseudo data $\{(\bx_t, \by_t^+)\}$ and $\{(\bx_t^+, \by_t)\}$ using models $\overrightarrow{p_n}^t$ and $\overleftarrow{p_n}^t$ respectively\\
        \nl Train $\overrightarrow{p_s}^t$ and $\overleftarrow{p_s}^t$ using $(\bx_t, \by_t^+)$ and $(\bx_t^+, \by_t)$\\
        }
        \nl \textbf{$//$ M-step:}\\
        \nl Generate denoised pseudo data \{$(\bx_t, \by_t^*)\}$ and $\{(\bx_t^*, \by_t)\}$ using $\overrightarrow{p_s}^t$ and $\overleftarrow{p_s}^t$\\
        \nl Train $\overrightarrow{p_n}^t$ and $\overleftarrow{p_n}^t$ using $\{(\bx_t, \by_t^*)\}$ and $\{(\bx_t^*, \by_t)\}$\\
        \nl Generate pseudo data $\{(\bx_t, \by_t^+)\}$ and $\{(\bx_t^+, \by_t)\}$ using $\overrightarrow{p_n}^t$ and $\overleftarrow{p_n}^t$ respectively\\
        \nl Train $\overrightarrow{p_n}^t$ and $\overleftarrow{p_n}^t$ using $\{(\bx_t^+, \by_t)\}\cup\{(\bx_t, \by_t^*)\}$ and $\{(\bx_t, \by_t^+)\}\cup\{(\bx_t^*, \by)\}$\\
        \nl $t := t + 1$\\      
        }
        \nl \textbf{return}  $\righttheta$, $\lefttheta$
\caption{Unsupervised NMT with SMT as PR} 
\label{alg:1}
\end{algorithm}

According to \newref{EM}, in the E-step, we need to minimize the gap between SMT models and NMT models. However, this step cannot be done by traditional gradient descent methods. Approximately, we train SMT models using the pseudo data generated by the corresponding NMT models to fit the mode of NMT posterior distributions. Thus the KL divergence between them is diminished. This step corresponds to the the $7^{th}$ and $8^{th}$ lines in Algorithm \ref{alg:1}, meaning SMT extracts good and frequent translation patterns from the data generated by current NMT models to finish denoising. 

In the M-step, we optimize two NMT models with gradient descent methods. We formulate the updating for $\lefttheta$ in \newref{params_updating}, to which that for $\righttheta$ is similar.

\begin{equation}
\begin{aligned}
&\nabla_{\lefttheta}\mathcal{J}(\righttheta,\lefttheta,\overrightarrow{p_s},\overleftarrow{p_s})\\
&= \mathbf{E}_{\bx\sim\overleftarrow{p_s}(\bx|\by_j)}\nabla_{\lefttheta}\log \overleftarrow{p_n}(\bx|\by_j;\lefttheta)\\
&+\mathbf{E}_{\by\sim\overrightarrow{p_n}(\by|\bx_i;\righttheta)}\nabla_{\lefttheta}\log \overleftarrow{p_n}(\bx_i|\by;\lefttheta)
\end{aligned}
\label{params_updating}
\end{equation}

\begin{table*}[ht]
\small
\begin{center}
\begin{tabular}{l|cccccc}
\hline
\multirow{2}*{Method}
 & \multirow{2}*{fr-en} & \multirow{2}*{en-fr} & de-en & en-de & de-en & en-de \\
 &  &  & (2014) & (2014) & (2016) & (2016) \\
\hline
\hline
\cite{artetxe2017unsupervised} & 15.56 & 15.13 & 10.21 & 6.89 & - & - \\
\cite{lample2017unsupervised} & 14.31 & 15.05 & - & - & 13.33 & 9.64 \\
\cite{yang2018unsupervised} & 15.58 & 16.97 & - & - & 14.62 & 10.86 \\
\cite{lample2018phrase}, NMT & 24.18 & 25.41 & - & - & 21.00 & 17.16 \\
\cite{lample2018phrase}, PBSMT & 27.16 & 28.11 & - & - & 22.68 & 17.77 \\
\cite{lample2018phrase}, NMT+PBSMT & 26.29 & 27.12 & - & - & 22.06 & 17.52 \\
\cite{lample2018phrase}, PBSMT+NMT & 27.68 & 27.60 & - & - & 25.19 & 20.23 \\
\hline
\textbf{Our Method} & 28.79 & 29.21 & 20.04 & 16.43 & 25.92 & 21.07 \\
\textbf{(+ R2L regularization)} & \textbf{28.92} & \textbf{29.53} & \textbf{20.43} & \textbf{16.97} & \textbf{26.32} & \textbf{21.65} \\
\hline
\end{tabular}
\end{center}
\caption{\label{tab:comparison} Comparison with previous methods.}
\end{table*}

This step corresponds to lines $14$ to $17$ in Algorithm \ref{alg:1}. A difficulty here is the exponential search space of the translation candidates. To address it, we leverage the sampling method \cite{shen2015minimum} and simply generate the top target sentence for approximation in our experiments. Note that in the 11$^{th}$ line, NMT models are trained using the denoised pseudo data generated by SMT models only, while in the 13$^{th}$ line, the mixed data of those and the pseudo data generated by the reverse NMT models are used. The intention here is to first use the denoised pseudo data to correct the NMT models established before, and then apply iterative back-translation to boost NMT models under the guide of the denoised data. NMT also makes up for the deficiency in smoothness of SMT in this step. In this way, SMT and NMT models can benefit from each other in the EM iterations.

\section{Experiments}
\subsection{Setup}
\subsubsection{Dataset}
In our experiments, we consider two language pairs, English-French and English-German. For each language, we use 50 million monolingual sentences in NewsCrawl, a monolingual dataset from WMT, which is the same as the previous work \cite{artetxe2017unsupervised,lample2018phrase}. For the convenience of comparison, we use \emph{newstest} 2014 as the test set for the English-French pair, and \emph{newstest} 2014 as well as \emph{newstest} 2016 for the English-German pair.
\subsubsection{Preprocess}
We use Moses scripts for word tokenization and truecasing. In model initialization, we use the public implementation of word2vec\footnote{https://github.com/tmikolov/word2vec} to train monolingual word embeddings of each language, and vecmap\footnote{https://github.com/artetxem/vecmap} to obtain cross-lingual embeddings of both language pairs. For NMT, we use the modified version of the public implementation\footnote{https://github.com/tensorflow/tensor2tensor} of Transformer \cite{vaswani2017attention}. We share the vocabulary space of 50,000 BPE codes \cite{sennrich2015neural} for source and target languages. For each language pair, we train two independent NMT models for different translation directions (i.e., source to target and target to source) with shared embedding layers of source and target sides. For SMT, we use the Moses implementation of PBSMT systems with Salm \cite{johnson2007improving}, which can denoise and reduce the size of phrase tables. And we use the default features defined in Moses for our PBSMT models. 

Our code is released in \url{https://github.com/Imagist-Shuo/UNMT-SPR}.

\subsection{Comparison}
\label{comparison}
\subsubsection{Baselines}
Our proposed method is compared with four baselines of unsupervised machine translation listed in the upper area of Table \ref{tab:comparison}, among which the fourth baseline contains several methods. Given a language pair, the first two baselines \cite{artetxe2017unsupervised,lample2017unsupervised} use a shared encoder and different decoders for the two languages. The third baseline \cite{yang2018unsupervised} uses different encoders and decoders, and introduces a weight sharing mechanism. The fourth baseline \cite{lample2018phrase} uses a shared encoder and decoder in their NMT systems. As for the training method, the second and third baselines use adversarial training. All of the four baselines use denoising auto-encoder and iterative back-translation.

Note that the fourth baseline contains four methods. ``NMT" means unsupervised NMT models, while ``PBSMT" denotes unsupervised SMT models with the back-translation method performed by SMT. ``NMT+PBSMT" and ``PBSMT+NMT" simply combine the best pseudo data that the former generates into the final iteration of the latter. Different from our proposed method, the training processes of NMT and SMT models in their methods are independent. 
 
\subsubsection{Results and Discussion}
The comparison results are reported in Table \ref{tab:comparison}. The BLEU scores are calculated by \emph{multi-bleu.pl}. From the table, we find that our method significantly outperforms all the baselines even the strong one \cite{lample2018phrase}. We elaborate the reasons as follows.

\textbf{(1)} Our proposed method significantly improves the performance over the ``NMT" and ``PBSMT" of \cite{lample2018phrase}. This is because unsupervised NMT methods suffer from the noise problem while PBSMT is inherently deficient in fluency just as the case study in \ref{case_study} shows. Our method can compensate for the deficiencies of them by combining the training processes of them. 
\textbf{(2)} Notice that ``NMT+PBSMT" performs even worse than pure ``PBSMT", which may be caused by accumulated errors in the iterations of NMT models. Due to the lack of timely denoising methods, infrequent errors and noises are repeated and reinforced as frequent ones by unsupervised NMT, so that even PBSMT could not distinguish them from good patterns in the last iteration. \textbf{(3)} The performance gained by ``PBSMT+NMT" verifies combining data of high quality into NMT training could be a better choice. But the simple combination in their method is not able to make the best of both models. In their method, NMT and SMT models are trained independently so that the bad patterns within the models themselves cannot be well removed due to weak supervision. In contrast, our proposed method integrates the training of NMT and SMT models in a unified EM framework where they can boost each other incrementally. The noises and errors generated by NMT models can be reduced in time by SMT as posterior regularization, while NMT can compensate for the deficiency of smoothness inherent in SMT models. Therefore, our proposed method still outperforms "PBSMT+NMT".

Apart from SMT as posterior regularization, our framework can be easily extended to incorporate other posterior regularization methods without changing model structures, such as the target-bidirectional agreement regularization \cite{Zhang2018RegularizingNM}. This regularization can help deal with the problem of exposure bias in supervised NMT, where another "reversed" NMT model is trained using data of reversed sentences from left to right. Then the "reversed" NMT model is leveraged to generate pseudo data for training the original NMT model. Specifically, we introduce the R2L regularization after the final training iteration of NMT models (i.e., NMT2 in Table \ref{tab:evolution}). With this extension, we achieve higher performance (+R2L regularization in Table \ref{tab:comparison}). 

\begin{table}[t!]
\small
\begin{center}
\begin{tabular}{l|cccc|c}
\hline
Steps & fr-en & en-fr & de-en & en-de & ave \\
\hline
\hline
E-step \ (SMT0) & 15.34 & 11.74 & 11.03 & 8.14 & 11.56\\
M-step (NMT0) & 24.06 & 24.82 & 16.29 & 12.88 & +7.95\\
\hline
E-step \ (SMT1) & 26.49 & 27.64 & 17.34 & 14.81 & +2.06\\
M-step (NMT1) & 28.29 & 29.02 & 19.61 & 16.02 & +1.67\\
\hline
E-step \ (SMT2) & 28.64 & \textbf{29.21} & 19.87 & 16.29 & +0.23\\
M-step (NMT2) & \textbf{28.79} & 29.17 & \textbf{20.04} & \textbf{16.43} & +0.11\\
\hline
\end{tabular}
\end{center}
\caption{\label{tab:evolution} Test BLEU on \emph{newstest} 2014 in different steps.}
\end{table}

\subsection{Model Evolution}
\label{evolution}
We conduct several EM iterations in our experiments, and record the test BLEU scores on \emph{newstest} 2014 after each E-step (SMT) and M-step (NMT) in Table \ref{tab:evolution}. We have tried more steps but the models do converge after three EM iterations. For the convenience of comparison, in the last column of the table, we also list the average improvement of four translation models after each step. From the table, first, we find NMT and SMT models improve incrementally after each iteration, which accords with our proposed motivation. Note that the improvements between adjacent NMT steps are exactly contributions made by SMT as posterior regularization. 
Second, the models improve the most in the first EM iteration and nearly converge at the third EM iteration. 

\begin{figure}[!htb]
\centering
\includegraphics[width=0.9\linewidth]{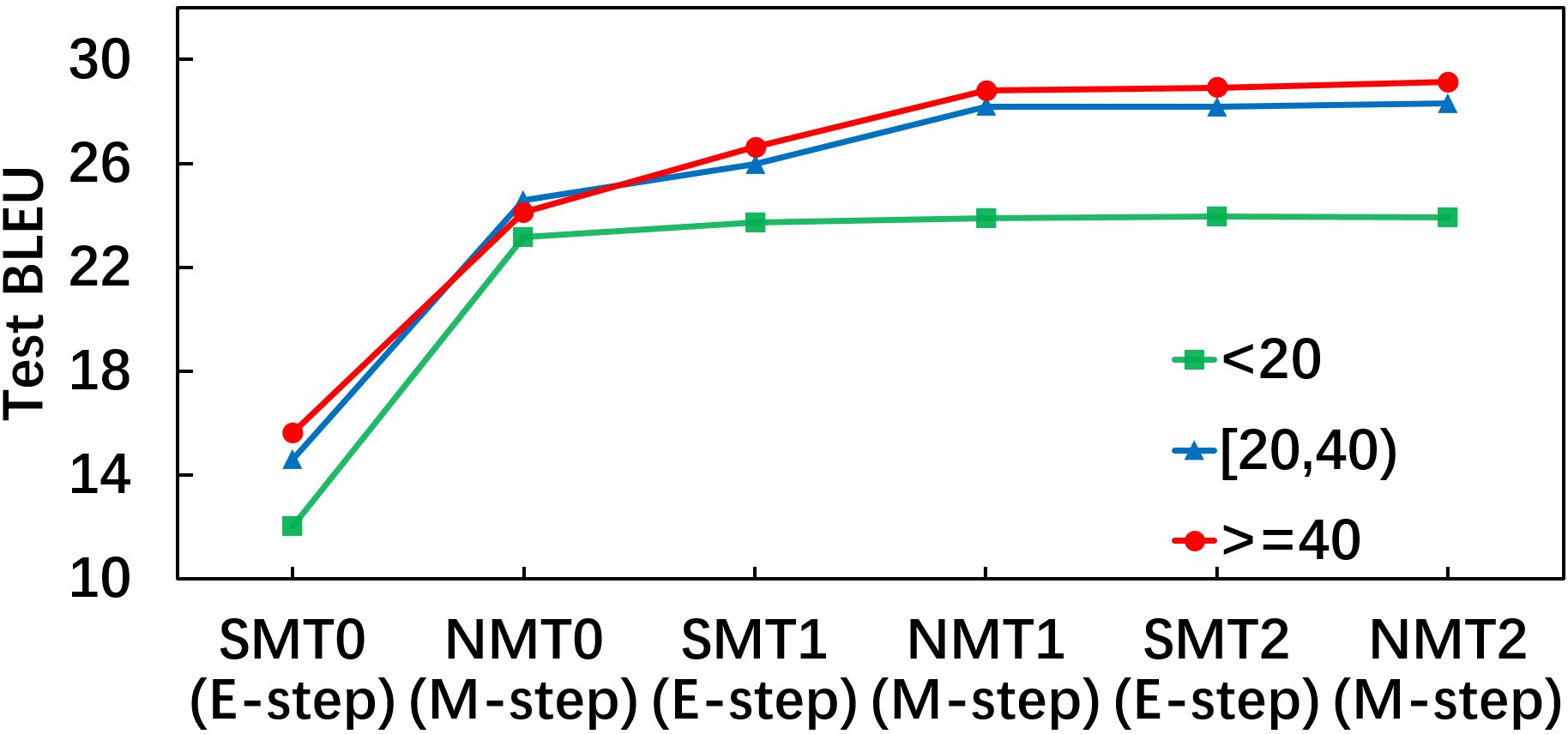}
\caption{Test BLEU on sentences grouped by length.}
\label{fig:evolution}
\end{figure}

Additionally, we also compare the translation performance on sentences of different lengths as iteration steps progress. We group the sentences in the fr-en test set by length as shown by the three curves in Figure \ref{fig:evolution}. Then, we record the BLEU scores of different groups after each step. From the figure, we find the models converge much slower on longer sentences, which indicates that it is easier for the models to learn shorter sentences. 

\subsection{Discussion on Initialization}
\begin{figure}[t]
\centering
\includegraphics[width=\linewidth]{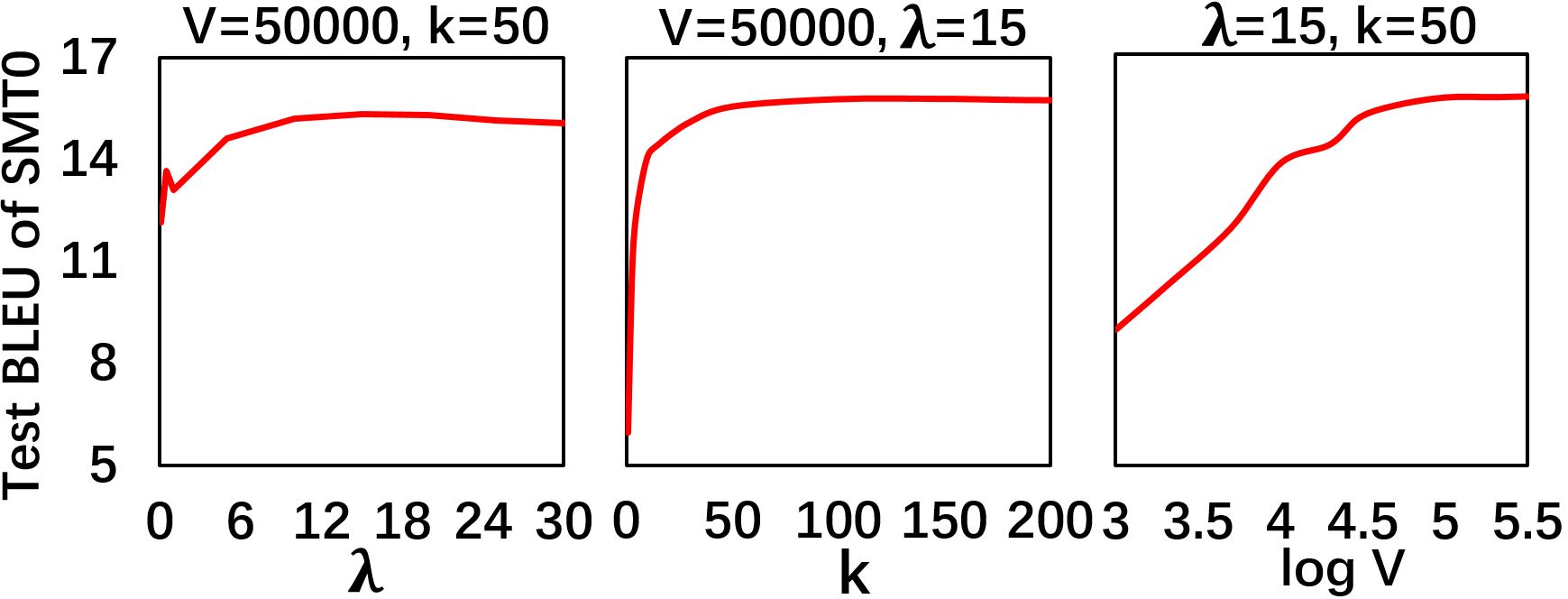}
\caption{Test of initial models with various hyper-params.}
\label{fig:init}
\end{figure}

In this subsection, we delve into the initialization stage which is crucial to our method.
In that stage, there are three hyper parameters described in \ref{sec_initialization} that should be taken into account, i.e., the peakiness controller $\lambda$, the vocabulary size $S$ or $T$, and the number of translation candidates $k$ for each word. Since the performance of initialization can be evaluated by SMT0, we adjust the hyper-parameters and measure the fr-en test BLEU of SMT0 models accordingly. For brevity, we let $S=T=V$ in our experiments. The results are illustrated in Figure \ref{fig:init}. From this figure, we find that $k$ and $V$ have much bigger impacts on the initial model SMT0 than $\lambda$. With the value of $\lambda$ increasing, the performance of SMT0 gradually improves but starts to decline a bit after around 20. This is because the larger $\lambda$ will make the distribution in \newref{word_translation_prob} sharper, severely restricting the search spaces of SMT models. Similarly, the performance of SMT0 improves in accord with the value of $k$ or $V$ going up. But the improvement stops after certain thresholds (about 80 of $k$ and 50000 of $V$). The reason may be the useful information provided by word-translation tables is saturated after those.  

We also tried other initialization methods in our experiments, such as directly using the pseudo parallel data constructed from word-by-word translation to warm up NMT models. We compare NMT0 models warmed up with this method (without SMT0) to NMT0 in our proposed method (with SMT0) in the following table, which stresses the necessity of SMT0 and the importance of good initialization.

\begin{table}[hp]
\small
\begin{center}
\begin{tabular}{l|cccc}
\hline
Initialization Method & fr-en & en-fr & de-en & en-de \\
\hline
\hline
NMT0 without SMT0 & 12.29 & 12.46 & 7.32 & 4.81\\
NMT0 with SMT0 & 24.06 & 24.82 & 16.29 & 12.88\\
\hline
\end{tabular}
\end{center}
\caption{\label{tab:init}The necessity of SMT0 in model initialization. The numbers in this table are BLEU scores on \emph{newstest} 2014.}
\end{table}

\begin{table*}[pth!]
\small
\begin{center}
\begin{tabular}{c|l}
\hline 
Source & J'ai eu des relations difficiles avec lui jusqu'\`{a} ce qu'il devienne vieux, \underline{malade}.\\
\hline 
SMT0 & I've gotten of difficult relations with him until he will become old, \underline{sick}.\\
NMT0 & I've had difficult relations with him until he's become old, \underline{ill-fated}.\\
SMT1 & I've had difficult relationships with him until he became old, \underline{sick}.\\
NMT1 & I had difficult relations with him until he became old and \underline{sick}.\\
\hline
Reference & I had a difficult relationship with him until he became old and \underline{ill}.\\
\hline\hline
Source & Le fonds d'investissement qui \'{e}tait propri\'{e}taire de cette \underline{b\^{a}tisse-l\`{a}} avait des choix \`{a} faire.\\
\hline
SMT0 & The owner of this underline{building}, so had to make a choice of which was an investment fund. \\
NMT0 & The investment fund that was an owner of that \underline{canopy-back business} had \underline{plenty of} choice to do.\\
SMT1 & The investment fund that was the owner of this \underline{building} just had to make choices.\\
NMT1 & The investment fund that was the owner of this \underline{building} had choices to make.\\
\hline
Reference & The investment fund that owned the \underline{building} had to make a choice.\\
\hline\hline
\multirow{2}*{Source} & M. Dutton a \underline{rendu visite \`{a}} Mme Plibersek pour garantir qu'aucun dollar \underline{du} plan de sauvetage ne sera d\'{e}pens\'{e}\\ &en bureaucratie suppl\'{e}mentaire.\\
\hline 
SMT0 &Mr Dutton \underline{paid a visit to} Ms Plibersek to guarantee that the greenback no rescue plan of not be spent in extra bureaucracy.\\
NMT0 & Mr Dutton \underline{said} Ms Plibersek'\underline{visit to} guarantee any dollar \underline{from} the rescue plan will be spent in extra bureaucracy. \\
SMT1 & Mr Dutton \underline{was visiting} Ms Plibersek to guarantee that no dollar rescue plan will be spent on additional bureaucracy.\\
NMT1 & Mr Dutton \underline{paid a visit to} Ms Plibersek to guarantee that no dollar \underline{from} the rescue plan will be spent on extra bureaucracy.\\
\hline
\multirow{2}*{Reference} & Mr Dutton \underline{called on} Ms Plibersek to guarantee that not one dollar \underline{out of} the rescue package would be spent on \\ &additional bureaucracy. \\
\hline
\end{tabular}
\end{center}
\caption{\label{tab:case} Cases of translation results from French to English in \emph{newstest} 2014. The models of SMT0, NMT0, SMT1 and NMT1 are corresponding to the steps in Table \ref{tab:evolution}.}
\end{table*}

\subsection{Case Study}
\label{case_study}
To further demonstrate the effectiveness of our method, we select some cases from translation results (fr-en) and compare the translations generated by models of different training steps. The results are listed in Table \ref{tab:case}. In the first case, which is exactly the example in the Introduction, the word ``malade" in French is wrongly translated into ``ill-fated" in English by NMT0. As we can see, this error has been corrected in NMT1 after the guidance of SMT1. In the second case, apart from the wrongly aligned word ``b\^{a}tisse-l\`{a}" to ``canopy-back business" by NMT1, there is also a redundant phrase ``plenty of" generated by it. Those errors are both corrected after the regularization of SMT1. In the third case, we also reach the same conclusion that NMT1 can benefit from SMT1 and rectify the mistake on ``rendu visite \`{a}".
There is also an interesting phenomenon from case three of NMT adhering to ``from" which makes the sentence more fluent, even though this word is missed by SMT models. In a word, the above analysis verifies that noises and errors in unsupervised NMT models can be eliminated timely by SMT models as posterior regularization with our method .

From these cases, we find that SMT can also benefit from NMT models. Even though the meanings of the key words could be captured by SMT, the outputs of SMT0 are not fluent especially in the second case. This problem is relieved in SMT1, after SMT is fed with more fluent pseudo data generated by NMT0, which validates that SMT and NMT can incrementally boost each other with our method.

\section{Related Work}

Previous unsupervised neural machine translation systems \cite{artetxe2017unsupervised,lample2017unsupervised,yang2018unsupervised} are mainly the modifications of the current encoder-decoder structure. To constrain outputs of encoders for two languages into a same latent space, \citeauthor{artetxe2017unsupervised} \shortcite{artetxe2017unsupervised}, and Lample et al. (2017) use a shared encoder, while \citeauthor{yang2018unsupervised} \shortcite{yang2018unsupervised} use a weight sharing mechanism. Denoising auto-encoder \cite{vincent2010stacked} and adversarial training methods are also leveraged to improve the ability of encoders. Besides, iterative back-translation is applied to generated pseudo parallel data for cross-lingual training.

After that, \citeauthor{lample2018phrase} \shortcite{lample2018phrase} summarize three principles for unsupervised machine translation, which are initialization, language modeling and iterative back-translation, and propose some effective methods with simplified training procedures. Four methods are leveraged in their work, including unsupervised NMT, unsupervised PBSMT and two combinations of them. Our method is different from them. In their methods, SMT and NMT are treated as independent models so that they suffer from respective deficiencies and cannot benefit from each other in their training processes. In contrast, we combine them into a unified EM training framework and enable them to improve jointly and boost each other incrementally, where NMT models are responsible for smoothing and fluency, while SMT models are responsible for denoising and guiding NMT models.

Moreover, there has been some work exploiting SMT features to improve supervised NMT. In \citeauthor{he2016improved} \shortcite{he2016improved}, the probability calculated by NMT is integrated as a feature into a log-linear model. After that, \citeauthor{tang2016neural} \shortcite{tang2016neural} and \citeauthor{wang2017neural} \shortcite{wang2017neural} leverage gate mechanisms to introduce a phrase table or candidates provided by SMT into NMT models. Different from them, we leave the model structures unchanged via the framework of posterior regularization. \citeauthor{zhang2017prior} \shortcite{zhang2017prior} also integrate more prior knowledge into the training of NMT with the help of posterior regularization. But there is a major difference that we introduce the successful practice of iterative back-translation into this framework with a unified EM training algorithm, where SMT and NMT models can benefit from each other. Additionally, in unsupervised scenarios, our SMT features are learned from scratch and improved incrementally, rather than pre-trained from real bilingual data and fixed during the whole procedure.

\section{Conclusion}
In this paper, we introduce SMT models as posterior regularization to denoise and guide unsupervised NMT models with the ability of constructing more reliable phrase tables and eliminating the infrequent and bad patterns generated in the back-translation iterations of NMT. We unify SMT and NMT models within the EM training algorithm where they can be trained jointly and benefit from each other incrementally. In the experiments conducted on en-fr and en-de language pairs, our method significantly outperforms previous methods, and achieves the new state-of-the-art performance of unsupervised machine translation, which demonstrates the effectiveness of our method. In the future, we may delve into the initialization stage, which is crucial to the final performance of the proposed method.

\bibliographystyle{aaai19}
\bibliography{aaai19}

\end{document}